\newcommand{\compileaaai}{}

\documentclass[letterpaper]{article} 
\usepackage{aaai24}  
\usepackage{times}  
\usepackage{helvet}  
\usepackage{courier}  
\usepackage[hyphens]{url}  
\usepackage{graphicx} 
\usepackage{natbib}  
\usepackage{caption} 
\usepackage{subcaption}

\usepackage[boxed,ruled,vlined,linesnumbered]{algorithm2e}
\SetAlCapFnt{\normalsize}
\SetAlgoInsideSkip{smallskip}
\SetAlFnt{\small}
\SetCommentSty{textit}
\SetInd{0.4em}{0.4em}

\usepackage{amssymb}
\usepackage{amsmath}
\usepackage{amsthm}
\usepackage{xspace}
\usepackage[dvipsnames]{xcolor}
\usepackage[textsize=scriptsize,textwidth=1cm]{todonotes}

\usepackage{ifthen}

\newcommand{\mytitle}{PRP Rebooted: Advancing the State of the Art in FOND Planning}

\title{\mytitle}

\author {
    Christian Muise\textsuperscript{\rm 1,3},
    Sheila A. McIlraith\textsuperscript{\rm 2,3},
    J. Christopher Beck\textsuperscript{\rm 2}
}
\affiliations {
    \textsuperscript{\rm 1} Queen's University, Kingston, Canada\\
    \textsuperscript{\rm 2} University of Toronto, Toronto, Canada\\
    \textsuperscript{\rm 3}
    Vector Institute for Artificial Intelligence, Toronto, Canada\\
    christian.muise@queensu.ca, sheila@cs.toronto.edu, jcb@mie.utoronto.ca
}

    \newcommand{\us}{PR2\xspace}

    \newcommand{\approximation}{reformulation\xspace}

\newcommand{\set}[1]{\lbrace #1 \rbrace}
\newcommand{\st}{\hspace{0.5em}|\hspace{0.5em}}

\newcommand{\act}{{\cal A}\xspace}

\newcommand{\pre}[1]{Pre_{#1}}
\newcommand{\eff}[1]{\mathit{Eff}_{#1}}
\newcommand{\vars}{\mathcal{V}}
\newcommand{\props}[1]{propositions(#1)}

\newcommand{\updateps}[2]{#1 \oplus #2}

\newcommand{\consistent}[2]{#1 \approx #2}

\newcommand{\fondregr}[3]{\mathcal{R}(#1, #2, #3)}
\newcommand{\fondproblem}[2]{\langle \vars, #1, #2, \act \rangle\xspace}

\newcommand{\regressable}{Can\mathcal{R}\xspace}

\newcommand{\psgraph}{\textsc{Controller}\xspace}
\newcommand{\csgraph}{\textsc{Reachable}\xspace}
\newcommand{\solstep}{SolStep\xspace}
\newcommand{\solsteps}{{\solstep}s\xspace}
\newcommand{\parents}{in}
\newcommand{\children}{out}

\newcommand{\fsap}[2]{\langle #1, #2 \rangle\xspace}

\newcommand{\hff}{$h^{FF}$\xspace}

\newcommand{\casesection}[1]{\vspace{0.5em}\noindent \textbf{#1}}

\theoremstyle{plain}
\newtheorem{theorem}{Theorem}

\theoremstyle{definition}

\newtheorem{definition}{Definition}

\newcommand{\myproof}[1]{\noindent\textit{Proof sketch.} {#1} \hfill $\square$\vspace{1mm}}

\newcommand{\myhide}[1]{}

\begin{document}

    \maketitle

\begin{abstract}
Fully Observable Non-Deterministic (FOND) planning is a variant of classical symbolic planning in which actions are nondeterministic, with an action's outcome known only upon execution. It is a popular planning paradigm with applications ranging from robot planning to dialogue-agent design and reactive synthesis. Over the last 20 years, 
a number of approaches to FOND planning have emerged. 
In this work, we establish a new state of the art, following in the footsteps of some of the most powerful FOND planners to date. Our planner, \us, decisively outperforms the four leading FOND planners, at times by a large margin, in 17 of 18
domains that represent a comprehensive benchmark suite.
Ablation studies demonstrate the impact of various techniques we introduce, with the largest improvement coming from our novel FOND-aware heuristic.
\end{abstract}

\section{Introduction}
Fully Observable Non-Deterministic (FOND) planning is a
variant of classical symbolic planning in which actions are non-deterministic, the finite set of possible action outcomes is known a priori, and the realized outcome in a particular instant is observed following execution of the action~(e.g.,  \citep{dan-tra-var-ecp99,cimatti2003}).
As with classical planning, FOND planning assumes full observability,
but the uncertainty in the outcome of actions during plan generation 
necessitates a new form of (contingent) solution -- both in terms of representation and plan~synthesis.

Since its introduction 20 years ago, FOND has emerged as a popular and highly versatile computational paradigm with applications ranging from generalized planning~\cite{ill-mci-aaai19,DBLP:conf/kr/BonetGGPR20}
and robot planning~\cite{AndresBD-FONDrobot-plan20} to dialogue-agent design~\cite{plan4dial}, and reactive synthesis from logical specification~\cite{camacho-ltl}.
Given the diversity of applications for FOND planning,
advances in FOND planning have the potential for significant impact.

Following the development of the original MBP FOND planner, based on model checking \cite{cimatti2003}, we
have witnessed a stream of innovations. In a departure from MBP, the NDP \cite{ndp} and FIP \cite{fip} planners reformulated the FOND problem into a classical planning problem, repeatedly solving this problem, to iteratively solve the original FOND problem. 
In 2012, the PRP planner emerged as the state of the art in FOND planning, often showing orders of magnitude improvements in plan generation time and/or solution size~\cite{prp}. Like NDP and FIP, PRP employed classical planning over a reformulation of the FOND problem, but much of its performance gains were the result of identifying and encoding only those aspects of the state that were relevant to plan validity. This was accomplished by representing families of states as conjunctive formulae and employing regression rewriting, a form of pre-image computation, to establish relevance~\cite{waldinger1977achieving,reiter2001knowledge,fritz2007}.
Subsequent techniques explored new concepts for computing solutions to FOND problems: from the backwards search of GRENDEL \cite{grendel}, to the SAT encoding of FONDSAT \cite{fondsat}, to the policy-based search of MyND and Paladinus \cite{mynd, paladinus}. These new planners were often published with additional benchmark problems that showcased the new planner's potential and, in a number of cases, their superior performance compared to the incumbent, PRP.

In this work, we introduce \textit{PRP Rebooted} (\us), a new FOND planner that leverages the insights of a generation of  FOND planners
to realize a significant advance in the state of the art. Like NDP, FIP, and PRP before it, \us adopts the approach of repeatedly replanning in a classical planning \approximation.
Further, it uses a powerful solution representation inspired by GRENDEL and FONDSAT, while maintaining the solving strength of PRP to produce this solution. The novel techniques introduced in the \us planner include better handling of deadends, a preprocessing step to simplify problems, and a new FOND-aware heuristic for the classical planning sub-process.

\us outperforms the four leading FOND planners (MyND, FONDSAT, PRP, Paladinus), at times by a large margin, in 17 of 18
domains that represent a comprehensive benchmark set (five of 800 problems are solved by PRP, but not \us).
Through an ablation study, we evaluate the impact of individual technical components on overall planner performance.
Ultimately, \us constitutes a significant advancement in the field of FOND planning and a new state of the art for the planning formalism.

\section{Preliminaries}
\label{sec:background}

Our notation and definitions largely follow the existing literature that uses multi-valued variables for representation.

\begin{definition}[FOND Task \citep{prp}]
A fully-observable non-deterministic (FOND) planning task is a tuple $\fondproblem{s_0}{s_*}$, where $\vars$ is a set of finite domain variables, $s_0$ is the initial state (complete setting of $\vars$), $s_*$ is the goal condition (partial assignment to $\vars$), and $\act$ is the set of potentially non-deterministic actions. Each action $a \in \act$ is represented by a tuple $\langle \pre{a}, \eff{a} \rangle$, where $\pre{a}$ is a partial assignment to $\vars$ that stipulates when the action is executable and $\eff{a}$ is a set of outcomes, one of which will occur at execution. Each outcome $o \in \eff{a}$ is a partial assignment to $\vars$ and signifies the updates to the state.
\end{definition}
For updating partial or complete states (represented as a partial or complete assignment to $\vars$ respectively), we define the $\updateps{}{}$ operator (where $p_1$ and $p_2$ are (potentially partial) assignments to $\vars$) as follows:
\begin{align*}
    (\updateps{p_1}{p_2})(v) = \begin{cases}
  p_2(v) & \text{ if $p_2(v)$ is defined} \\
  p_1(v) & \text{ otherwise}
\end{cases}
\end{align*}

We say that partial states $p_1$ and $p_2$ are \textit{consistent} (written $\consistent{p_1}{p_2}$) when $\forall v \in \vars$, either $p_1(v) = \bot$ or $p_2(v) = \bot$ or $p_1(v) = p_2(v)$, where $\bot$ indicates the variable is undefined. Action $a \in \act$ can be executed in state $s$ only when $\consistent{s}{\pre{a}}$. If outcome $o \in \eff{a}$ occurs as a result of applying $a$ in state $s$, then the updated state is defined as $\updateps{s}{o}$.

In FOND, an initial state $s_0$ and set of actions ${\cal A}$ induce a set of reachable states. Despite the nondeterminism in FOND actions, a policy, $\pi$, is a mapping from  states to individual actions (rather than an action distribution, for example). We restrict our attention to defining this mapping for the subset of states that are reachable from $s_0$, and as such we refer to such policies as a \emph{partial policy}.
If every state reachable by a policy itself has a mapping to an action
then we say that the policy is \textit{closed}.
Here, a \emph{plan} is a policy that is 
guaranteed to achieve a goal, potentially under some assumptions or restrictions.
In the context of FOND planning, a \textit{weak} plan is one that will reach the goal under some realization of the non-deterministic action effects; it need not be, and is typically not, closed.
A \textit{strong} plan is a closed (partial) policy and is guaranteed to reach the goal in a finite number of steps.
Finally, a \textit{strong cyclic} plan is a closed (partial) policy where the policy embodies a weak plan for every state that is reachable by the partial policy. A strong cyclic plan provides a solution to the FOND planning under an assumption of fairness \citep{cimatti2003}.
In this paper, we are concerned with strong cyclic plans.

As is common with several FOND approaches, we use an \textit{all-outcomes determinization}. This is a reformulation of the problem so that every action $a \in \act$ is replaced with $|\eff{a}|$ actions, each corresponding to one of the outcomes in $\eff{a}$. Solving the classical planning problem created by the all-outcomes determinization gives us a sequence of actions that achieves the goal under some possible realization of environmental uncertainty and, thus, is a weak plan.
Much of our work relies on
computing relevant conditions for an action to achieve some partial state. We exploit regression, a pre-image computation~\cite{waldinger1977achieving,reiter2001knowledge}.

\begin{definition}[Logical Regression]
Given a partial state $p$, action $a \in \act$, and outcome $o \in \eff{a}$, the logical regression of $p$ through the action $a$ and outcome $o$ is defined as:
\begin{align*}
    \fondregr{p}{a}{o}(v) = \begin{cases}
        \pre{a}(v) & \text{ if $\pre{a}(v) \neq \bot$} \\
        \bot & \text{ else if $p(v) = o(v)$} \\
        p(v) & \text{ otherwise}
    \end{cases}
\end{align*}
We only define (and apply) regression if $\consistent{p}{o}$.
We use $\regressable(p)$ to denote all of the actions and outcomes that have regression defined for the partial state $p$:
\begin{align*}
    \regressable(p) = \lbrace \langle a, o \rangle \st &a \in \act, o \in \eff{a}, \text{and } \consistent{p}{o}\rbrace.
\end{align*}
\end{definition}

Finally, modern FOND planners need to avoid deadends -- reachable states from which the goal is not reachable. We adopt PRP's concept of a \textit{forbidden state-action pair} (FSAP). An FSAP is a tuple $\langle ps, a \rangle$, denoting that action $a$ is forbidden in any state consistent with partial state $ps$ because it could lead to a deadend.
In the algorithms that follow, $\mathit{fsap} = \langle ps, a \rangle$ and we use $\mathit{fsap}.ps$ and $\mathit{fsap}.a$ to reference the corresponding partial state and associated action.

\section{Approach}
\label{sec:approach}
Similar to NDP, FIP, and PRP, \us iteratively solves a determinized version of the FOND planning problem to compute weak plans, incorporating them into an overall solution. We view \us as an evolution of PRP, 
as we use a similar high-level search framework, leveraging regression rewriting for relevance determination and forbidden state-action pairs for deadend avoidance. The \us planner is also built on top of the Fast Downward planning system \cite{fd} but otherwise shares minimal implementation overlap with PRP.

The overall approach is summarized in Algorithm \ref{alg:high-level} and described throughout this section 
\ifthenelse{\equal{\compileaaai}{}}%
{(see Appendix \ref{app:high-level} for an expanded version).}%
{(cf. Appendix A in \cite{pr2arxiv} for an expanded version).}
\us incrementally builds a solution, potentially making missteps along the way. If the found solution is strong cyclic, then we are done. Otherwise, information is retained on what went wrong, and the process restarts.

\begin{algorithm}[t]
\KwIn{FOND planning task, $\Pi = \fondproblem{s_0}{s_*}$}
\KwOut{Policy}
incumbent = \texttt{make\_empty\_solution}(); \hspace{1em} FSAPS = $\emptyset$\;
\While{!incumbent.is\_strong\_cyclic()}
{
  sol = \texttt{make\_empty\_solution}($\set{s_0}$)\;
  \While{sol.\csgraph contains unhandled nodes \label{line:while-openlist}}
  {
    n = sol.\csgraph.pop\_unhandled\_node()\;

    \vspace{0.5em}
    \Switch{\texttt{analyze\_node} (n)}{
        \textbf{case 0:} \texttt{skip\_if\_strong\_cyclic}(n)\\
        \textbf{case 1:} \texttt{skip\_if\_poisoned}(n)\\
        \textbf{case 2:} \texttt{match\_complete\_state}(n)\\
        \textbf{case 3:} \texttt{apply\_predefined\_path}(n)\\
        \textbf{case 4:} \texttt{match\_complete\_state}(n)\\
        \textbf{case 5:} \texttt{find\_and\_update\_weak\_plan}(n)\\
        \Case{default (\textbf{case 6})}{ 
            \texttt{record\_deadend}(n)\;
            \If {n.state == $s_0$}
            {
                \Return{make\_policy(incumbent.\psgraph)}\;
            }
        }
    }
  }
  \If {sol.success\_rate() $\geq$ incumbent.success\_rate()}
  {
    incumbent = sol\;
  }
}

\Return{make\_policy(incumbent.\psgraph)}\;
\caption{\us High-Level Planner}
\label{alg:high-level}
\end{algorithm}

\subsection{Solution Representation and Construction}

Algorithm \ref{alg:high-level} endeavours to compute a strong cyclic plan by incrementally building a solution from weak plans in the all-outcomes determinization. A key component of \us is the internal representation of the incumbent solution in terms of two structures: (1) a controller - a directed graph whose nodes capture (among other things) a compact representation of state-action pairs (i.e., a state the solution can  
reach and an action that is applicable in that state), and outgoing edges connecting to successor 
nodes; and (2) the reachable state space explored by the solution to this point, where nodes are \textit{complete} state-action pairs.
We refer to the first as \psgraph and the second as \csgraph (i.e., an incumbent solution $sol = \langle \psgraph, \csgraph \rangle$), with the former inspired by GRENDEL and FONDSAT and the latter similar to MyND, FIP, and Paladinus. Both play an essential role to \us:
the dual representation allows \us to maintain the inner-loop search progress with the reachable state space while maintaining a compact controller representation of the partial solution.
Intuitively, \csgraph represents the search progress of \us, including the open nodes that must be extended for a final solution, while \psgraph is a compact representation of a partial solution, including the conditions under which it is guaranteed to achieve the goal. For both graphs, we use $outcome(n_1, n_2)$ to label the edge between $n_1$ and $n_2$ with 
the outcome -- from the set of effects of the action of $n_1$ in this context.
More formally, the nodes in \psgraph are as follows.

\begin{figure}[t]
    \centering
    \includegraphics[width=\columnwidth]{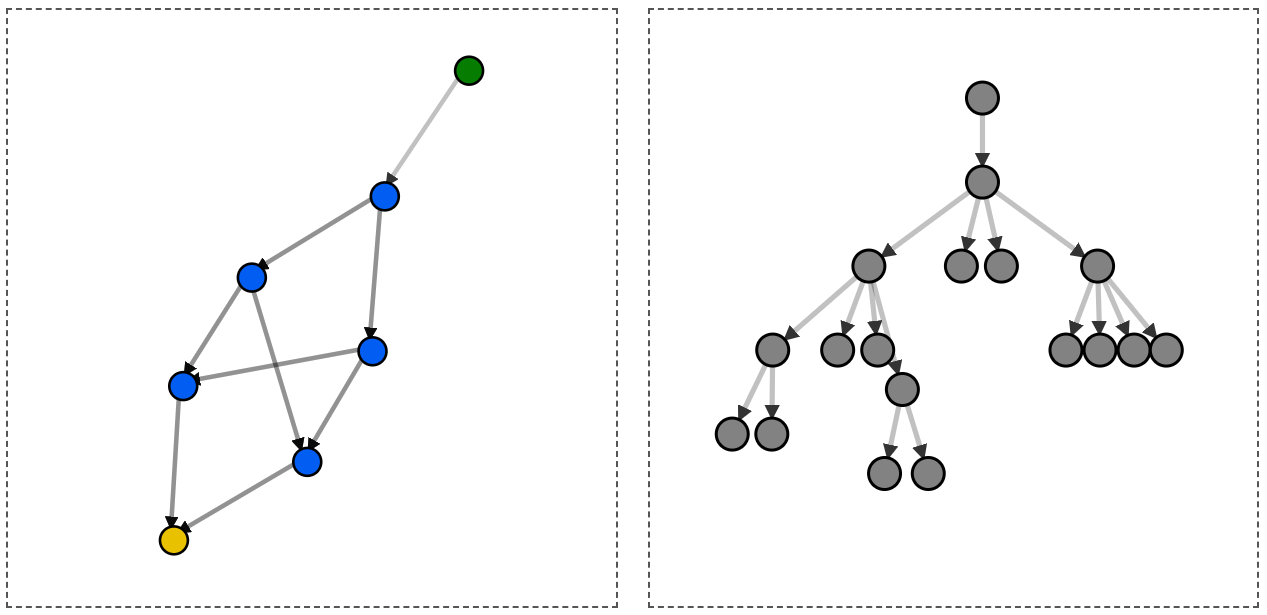}
    \caption{\psgraph (left) and \csgraph (right) representations. In the \psgraph, green represents the initial \solstep to execute, gold the goal, and blue corresponds to \solsteps that have been established as strong cyclic (and thus a solution is guaranteed from that point on).}
    \label{fig:tt-policy-example}
\end{figure}

\begin{definition}[\solstep]
A \solstep, $ss$, is a node in the \psgraph represented by a tuple, $ss = \langle ps, a, \parents, \children, sc \rangle$ where $ps$ is a partial state, $a$ is an action, $\parents$ is a list of \solsteps that are connected to $ss$ through an incoming edge (i.e., predecessors), $\children$ is the list of successors (directly corresponding to the outcomes of $a$), and $sc$ is a Boolean flag that indicates if the \solstep is ``strong cyclic'' (discussed below). Every \solstep in (the possibly empty) $ss.\parents$ is defined, but some in $ss.\children$ may be undefined.
\end{definition}

In Figure \ref{fig:tt-policy-example}, we show the \psgraph and \csgraph representations. 
We may not explore every complete state reachable by the solution -- unlike planners that rely solely on representations like \csgraph\ -- as we only continue exploring if \psgraph is not guaranteed to succeed.
While a single node in \csgraph corresponds to exactly one \solstep (i.e., node in \psgraph), it is often the case that a single \solstep corresponds to several nodes in \csgraph. We use $\csgraph(ss)$ to refer to the set of nodes in \csgraph that are associated with \solstep $ss$, and $\psgraph(n)$ to refer to the single \solstep in \psgraph that is associated with node $n$ in \csgraph.

Algorithm \ref{alg:high-level} determines how a node in the search space should be handled via a single case \texttt{analyze\_node} (line 6) and a final catch-all case for nodes that are deadends (line 13).
Next, we summarize how each case is identified and maintains the two components of our incumbent solution.

\casesection{Case 0: Strong Cyclic Nodes}
As a base case, if the search node we pop from the open list is handled by a node in \psgraph that is strong cyclic, then nothing need be done.

\casesection{Case 1: Poison Nodes}
A node is ``poisoned'' if it, or one of its ancestor in \psgraph, is flagged as a forbidden state-action pair (FSAP). In this situation, we do not process the node further and the search continues. The poisoning of nodes happens in Case 6 with details in Section \ref{sec:deadends}.

\casesection{Case 2: Complete State Match}
In this case, we have an exact match of the state that corresponds to the popped node and a 
node in \csgraph. At this point, we can assume that the search for a solution from this state has already run its course, and we connect things appropriately in both \psgraph and \csgraph. This may involve creating a new edge in \psgraph but not introducing new nodes.

\casesection{Case 3: Predefined Path}
In this case, there is already a predefined path for the search node: the \solstep associated with the parent search node in \psgraph has an edge defined for the outcome that has led to this particular search node. We, therefore, simply add all of the successor nodes to \csgraph according to the action specified in \psgraph. This situation arises when the \csgraph being constructed stumbles upon a part of the \psgraph solution that already contains enough information to handle those newly reached complete states. It is effectively overlaying new complete states in the expanded \csgraph on top of existing \solsteps in \psgraph. If the \psgraph nodes were marked strong cyclic, then this node would have been handled with Case 0.

\casesection{Case 4: Hookup Solution Steps}
Case 4 arises when the complete state corresponding to the current search node can be handled by a \solstep in \psgraph. Formally, for node $n$ and \solstep $ss$, this means that $\forall v \in \vars$, if $ss.ps \neq \bot$ then $n.state(v) = ss.ps(v)$. When this holds, a new edge is created from the previous search node's \solstep to this new matching one along the corresponding outcome.
Case 4 can be seen as a generalization of Case 2.
When these new connections are made, the \psgraph is updated through the fixed-point regression procedure defined in Section \ref{sec:fpr} that ensures that the partial states associated with each \solstep in \psgraph capture precisely what must hold in order for the solution to be strong cyclic from that point on.

\begin{figure}[t]
    \centering
    \includegraphics[width=\columnwidth]{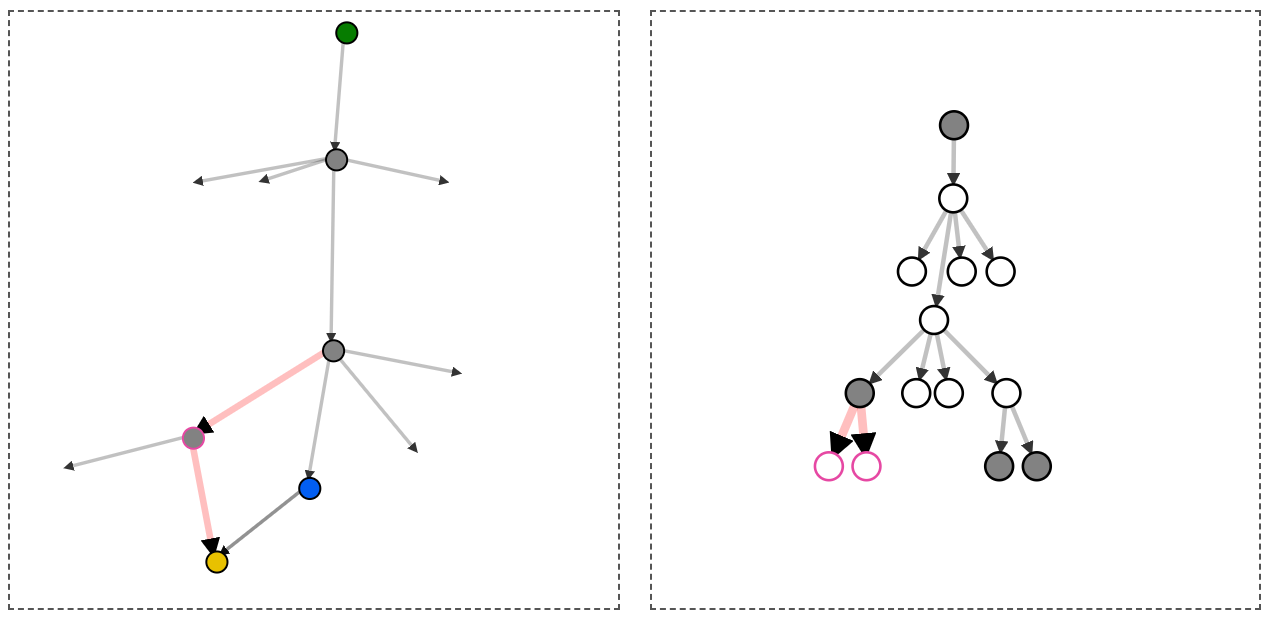}
    \caption{New path corresponding to a weak plan in the \psgraph (left) and \csgraph (right). In the \psgraph, highlighted edges show the new path, and dangling edges correspond to parts of the solution yet to be explored. In the \csgraph, the newly added nodes are highlighted, and white nodes correspond to search nodes that are still on the stack in Algorithm \ref{alg:high-level}.}
    \label{fig:newpath}
\end{figure}

\casesection{Case 5: Find and Update Weak Plan}
If the previous cases fail to capture the current search node (and associated state), we turn to finding another weak plan. We take the state associated with the current node in the search and first attempt to ``plan locally'' before computing a new plan for the goal. Planning locally is a strategy adapted from PRP, where the goal is temporarily set to the partial state of the \solstep in \psgraph that the parent \solstep expected to be in -- every time a weak plan is produced, one outcome is chosen in the all-outcomes determinization, and this dictates the temporary goal for planning locally.
Figure \ref{fig:newpath} illustrates a new (short) path that was found for a Case-5 node.
Note that \texttt{find\_and\_update\_weak\_plan} will add new nodes and edges to both the \psgraph and \csgraph graph. It also adds all of the successors of node $n$ to the \csgraph so that they may be subsequently considered.
Our weak planning procedure also includes (1) stopping the search when a state is reached such that some \solstep in the \psgraph matches (and thus we have a solution); (2) recording and generalizing all deadends found (cf. Section \ref{sec:deadends}); and (3) using an FSAP-aware heuristic that takes non-determinism into account (cf. Section \ref{sec:heuristic}).

\casesection{Case 6: Deadend}
In the final case, there is no weak plan to take us to the goal from the current node's state. We flag the current state as a deadend, generalize and apply regression if possible, and then poison the parent of this node and all of the parent's descendants in \csgraph (cf. Section \ref{sec:deadends}).

\smallskip The final task of Algorithm \ref{alg:high-level} (line 19) is to convert the solution's \psgraph into a policy. For a complete state $s$, assume $ss$ is the \solstep closest to the goal among all consistent \solsteps  (i.e., $\consistent{s}{ss.p}$). The action executed by \textit{make\_policy}(\psgraph) for state $s$ would thus be $ss.a$.

\subsection{Strengthening \& Fixed-Point Regression}
\label{sec:fpr}

Ideally, our \psgraph embodies the conditions required for a solution fragment to achieve the goal. Specifically, each node $n$ in \psgraph has a partial state associated with it that captures \textit{what must be true in order for the \psgraph to be a strong cyclic solution for a complete state that is consistent with $n.p$}. When a new connection is made in the \psgraph, we must apply regression on that connection. This, in turn, may alter the partial state associated with the new connection's source node, which then necessitates the repeated regression recursively back through the \psgraph until no further updates are made.

\begin{figure}[t]
    \centering
    \includegraphics[width=0.9\columnwidth]{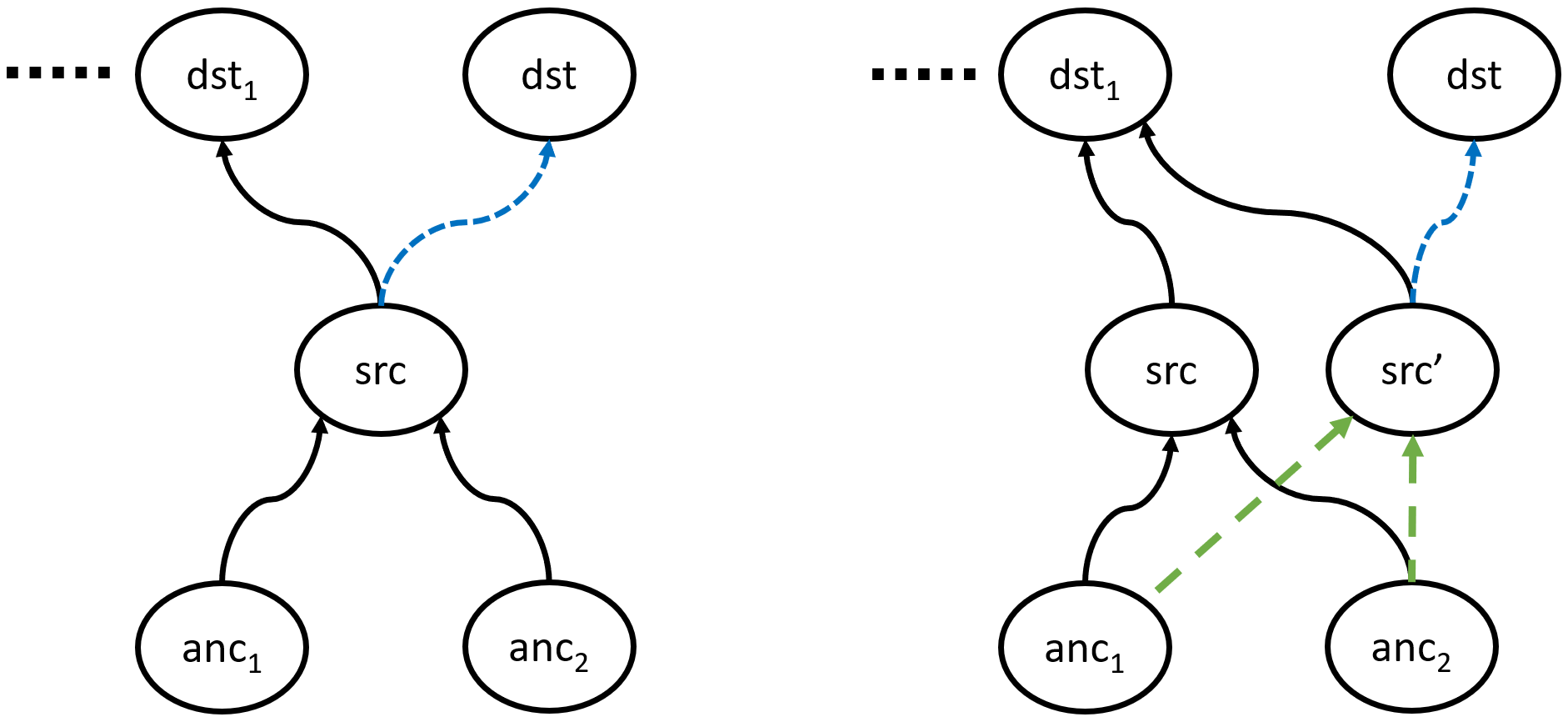}
    \caption{New connection in Fixed Point Regression}
    \label{fig:fpr}
\end{figure}

Figure \ref{fig:fpr} shows the situation where we have a new {\color{blue} blue} connection being made in \psgraph between \solsteps $src$ and $dst$ (assuming $dst$ is not already in $src.\children$). The recursive call is made with $src$ and $dst$, and is shown in Algorithm \ref{alg:fpr}. The {\color{blue} blue} edge between $src'$ and $dst$ is the newly formed link, and the {\color{ForestGreen} green} edges between $anc_1$ ($anc_2$) and $src'$ show the recursive calls made on Algorithm \ref{alg:fpr}'s line \ref{line:fpr:recurse}.

\begin{algorithm}[t]
\KwIn{Two nodes $src$ and $dst$ in \psgraph that should be connected, along with the associated \csgraph nodes $n_{src}$ and $n_{dst}$.}

\vspace{0.5em}
\tcp{If either $src$ or $dst$ are poisoned, then return since this part of the \psgraph will be discarded.}
\If{\texttt{is\_poisoned($src$)} or \texttt{is\_poisoned($dst$)}}
\Return{}

\vspace{0.5em}
\tcp{Compute the combined regression}
$ps = \updateps{src.p}{\fondregr{dst.p}{src.a}{outcome(src,dst)}}$\;

\vspace{0.5em}
\If{$ps == src.p$}
{\Return \hspace{1em} \tcp{Nothing to update}}

\vspace{0.5em}
\tcp{Clone $src$ and add it to \psgraph}
$src' = \langle ps, src.a, src.\parents, src.\children \cup \set{dst}, src.sc \rangle$\;
\psgraph.\texttt{add\_node($src'$)}\;

\vspace{0.5em}
\tcp{Update the graph mappings}
\psgraph($n_{src}$) = $src'$\;
\csgraph($src$) = \csgraph($src$) $\setminus$ $\set{n_{src}}$\;
\csgraph($src'$) = $\set{n_{src}}$\;

\vspace{0.5em}
\tcp{Recurse backwards}
\For{$anc \in n_{src}.\parents$}
{
    \texttt{fpr($\psgraph(anc)$, $src'$, $anc$, $n_{src}$)}\;\label{line:fpr:recurse}
}
\caption{Fixed-Point Regression, \texttt{fpr}}
\label{alg:fpr}
\end{algorithm}

The algorithm repeatedly strengthens the conditions on the nodes found in \psgraph, and leaves the strengthened version as a new copy. We create new copies rather than overwrite the original since different paths in \csgraph may overlay the same nodes in \psgraph and there is no guarantee that the states on the other paths will be consistent with the newly computed partial states.

Due to space, we forgo elaborating on the finer details of the process, including optimizations for when we can avoid cloning $src$, bookkeeping that removes unused aspects of the \psgraph, and proper maintenance of the strong cyclic property on nodes in \psgraph.

\subsection{Full Strong Cyclic Marking}
\label{sec:scm}

Because of the complex ways modifications are made to the \psgraph, we re-establish the strong cyclic property of the \psgraph nodes every time there is a modification that may change their status. The property captures a guarantee that the goal will be reached if the state of the world matches the partial state associated with a node in the \psgraph marked ``strong cyclic''.
The procedure for re-establishing the strong cyclic property of all nodes is given in Algorithm \ref{alg:scm}, and it assumes \solsteps in \psgraph default to having the $.sc$ property be false.

\begin{algorithm}[t]
\KwIn{Incumbent \psgraph}

\vspace{0.5em}
\tcp{Identify all of the nodes not marked strong cyclic}
$unmarked = \set{ n \st n \in \psgraph \text{ and } \neg n.sc}$\;

\vspace{0.5em}
\tcp{Flag not strong cyclic those with an unhandled successor}
$notSC = \emptyset$\;
$Q = \emptyset$\;
\For{$n \in unmarked$}
{
    \If{$|n.out| \neq |\eff{n.a}|$}
    {
        $notSC.add(n)$\;
        $Q.add(n)$\;
    }
}

\vspace{0.5em}
\tcp{Recurse backwards, marking more non-strong cyclic}
\While{$Q$ is not empty}
{
    $n = Q.pop()$\;
    \For{$n' \in n.in$}
    {
        \If{$n' \notin notSC$}
        {
            $notSC.add(n')$\;
            $Q.add(n')$\;
        }
    }
}

\vspace{0.5em}
\tcp{Finally, mark all remaining ones as strong cyclic}
\For{$n \in (unmarked \setminus notSC)$\label{line:scm:final-for}}
{
    $n.sc = true$\;
}

\caption{Strong Cyclic Marking}
\label{alg:scm}
\end{algorithm}

The algorithm relies on a few key assumptions. One is that the conditions associated with a \solstep are sufficient for executing from that point on, regardless of future action outcomes. This is guaranteed by the fixed-point regression procedure discussed in Section \ref{sec:fpr}. The other property is that the $.sc$ property of a \solstep is monotonic, in the sense that once it becomes true, it will stay true during the construction of \psgraph.

Intuitively, this is a safe assumption to make because of the manner in which the \psgraph is constructed -- a \solstep is added only if there is a path to the goal, and no modification removes this property. 
Seen another way, if there is some sequence of actions and outcomes that follows from a \solstep in \psgraph and leads to an unhandled outcome, then that final \solstep $n$ would necessarily have $n.sc = false$. Inductively, this property would apply all the way back through its predecessors. The nodes that remain in $(unmarked \setminus notSC)$ on line \ref{line:scm:final-for} must then have a path to the goal down every possible extension and can safely be marked as strong cyclic.

\subsection{Deadend Handling \& Poisoning}
\label{sec:deadends}

If there is any point during the search when we discover a deadend as part of the solution, we take an aggressive strategy to immediately invalidate any aspect of the search that is impacted. The strategy has two components: (1) deadend generalization and regression and (2) node poisoning.

We first employ deadend generalization in the same way as SixthSense and  PRP \cite{KolobovMW10,prp}: variables are progressively relaxed as long as a delete-relaxed \cite{DBLP:journals/ai/BonetG01} deadend remains, and the final partial state represents a generalization of the complete-state deadend that was detected. Similar to PRP, the generalized deadend $de$ is then regressed through every action and outcome that is consistent with $de$, in order to produce a set of forbidden state-action pairs (FSAPs) used in subsequent search iterations:
\begin{align*}
FSAPs(de) = \set{\fsap{\fondregr{de}{a}{o}}{a} \st \langle a,o \rangle \in \regressable(de)}.
\end{align*}

Second, given the node $n$ that was determined to be a deadend, we poison the portion of the \csgraph search space corresponding to the parent of $n$ and all the parent's descendants. As described in Case 6 above, a poisoned node is subsequently ignored in this iteration. This reasoning follows from how the top-level search proceeds: if we cannot guarantee the goal is reachable from a node, then future passes should skip this part of the search entirely.  \us stops looking for a solution down a path that will be skipped in a future iteration, thus saving search effort.

\subsection{FSAP-aware FF Heuristic}
\label{sec:heuristic}

FOND planners based on weak planning procedures carry very little of the FOND setting to the classical planner sub-calls \cite{paladinus}. \us addresses this issue in several ways. Similar to PRP, \us (1) stops searching when the incumbent solution recognizes the state, and (2) does not consider forbidden actions when expanding a search node.
\us further employs a custom classical heuristic that uses information from the FSAPs found so far to re-weight the \hff heuristic computation \cite{ff}.

The \hff heuristic operates by finding a plan in the \textit{delete relaxation} of the classical planning problem -- all delete effects are ignored (in our case, a variable can take on multiple values). The actions at the start of this delete-relaxed plan are called \textit{helpful actions}. Two key changes were made to the FF heuristic to take the FSAPs into account. First, when computing the helpful actions for a state, we only allow those that are not forbidden.
Second, we change the heuristic value by adding a penalty to any potentially forbidden action that is used in the computation of the heuristic. This ``potentially forbidden'' property is calculated by observing which FSAPs have their conditions satisfied by the propositions reachable in the delete relaxation.
Such actions may not truly be forbidden by the time they are needed/used. Thus, removing them entirely may cause the state to be mistakenly presumed to be a deadend. Thus, we only treat their presence with a penalty to the heuristic value (i.e., ``pay a price'' for needing an action that may potentially be forbidden).
\ifthenelse{\equal{\compileaaai}{}}%
{See Appendix \ref{app:heuristic} for further details.}%
{See Appendix B in \cite{pr2arxiv} for further details.}

\subsection{Redundant Object Sampling}
\label{sec:obj-sampling}

Prior to attempting to solve a problem, \us will explore simplifications to the instance 
by attempting to remove redundant objects.\footnote{We forgo introducing objects in the paper, as it is only this one contribution that makes use of them. Intuitively, actions are parameterized by objects, allowing us to represent the domain specification compactly. Cf.
\cite{pddlbook} for more details.} This sound-but-incomplete transformation comes from the following observation: \textit{if we delete some of the objects in a problem, and the resulting instance has a strong cyclic plan, then it must be a plan to the original problem.} Intuitively, removing objects has an impact equivalent to disallowing or deleting several actions, thus not invalidating any plans that remain -- all remaining (ground) actions exist identically in the unmodified domain. The goal remains unchanged as well. Therefore, a solution to the simplified problem is also one for the original problem.
This is a simple pre-processing step that could, in theory, be used with any FOND planner. Further details on the object sampling and the hyperparameters chosen (i.e., how many objects to remove and how long to search for a solution) can be found in
\ifthenelse{\equal{\compileaaai}{}}%
{Appendix \ref{app:object-sampling}.}%
{Appendix C of \cite{pr2arxiv}.}

\subsection{Force 1-safe Weak Plans}
\label{sec:1safe}
When a weak plan is found (cf. Case 5 of Algorithm \ref{alg:high-level}), rather than immediately adding the newly found nodes to both \psgraph and \csgraph, we confirm that no immediate extension leads to a deadend. This is a simple process that involves (1) computing every state reachable from the actions in the plan (which will eventually become open nodes in \csgraph); (2) checking if any such states are deadends; (3) recording the relevant generalized deadends and FSAPs, and then re-running the weak planning procedure if a deadend is detected.
We are guaranteed to find a different weak plan if step (3) is taken, since the weak planning procedure will avoid the deadend discovered in step (2) due to new FSAPs. This optimization can potentially lead to far fewer updates to the \psgraph and \csgraph.

\section{Theoretical Properties}

The non-deterministic planning techniques exploited in \us yield a number of interesting theoretical properties. In what follows, we provide the proof sketches for two of the most important results: the soundness and completeness of the \us planner.

\begin{theorem}[Soundness]
For a given FOND task $T$, if a strong cyclic plan \psgraph is produced by \us, then it is a strong cyclic plan for $T$.
\end{theorem}

\myproof{The soundness of the approach rests on the \psgraph representation, and what it means for $incumbent.is\_strong\_cyclic()$ to return true in Algorithm \ref{alg:high-level}. This check amounts to inspecting the $sc$ property of the \solstep that corresponds to the initial state, and Algorithms \ref{alg:fpr} + \ref{alg:scm} combine to ensure that this initial \solstep is marked strong cyclic only if the full \psgraph is.

Note that in Algorithm \ref{alg:scm}, nodes that are marked on the final line will be those that (1) have all their successors defined (cf. the first for-loop) and (2) cannot reach some unmarked node (cf. the second for-loop). Because every node in \psgraph marked $sc$ cannot reach an unmarked node, we can conclude that a node newly marked $sc$ can always reach the goal under the assumption of fairness.

The final step is to consider the continued executability of actions used when following \psgraph. By using the combined regression in Algorithm \ref{alg:fpr}, we are guaranteed that for any \solstep $ss$, executing $ss.a$ in a state $s$ where $s \models ss.ps$, the outcome taken will lead to a state $s'$ that corresponds to the successor \solstep. This invariant on the \psgraph that ties together neighbouring \solsteps is the same one that leads to the soundness of the FONDSAT and GRENDEL planners.}

\begin{theorem}[Completeness]
For a given FOND task $T$, if a strong cyclic plan exists, then \us will eventually find one.
\end{theorem}

\myproof{Completeness intuitively follows from the eventual discovery of all required FSAPs. If \csgraph leads to a place in the search space where no strong cyclic solution exists, then a deadend will be discovered and new FSAPs created. The outer-loop of Algorithm \ref{alg:high-level} will repeat as long as new FSAPs are discovered, and so we need only consider the final round (assuming the FSAPs found are correct and only finitely many may exist -- a natural consequence of the finite nature of the domain).

The correctness of the identified FSAPs (cf. Section \ref{sec:deadends}) is established inductively by the correctness of the deadends they lead to. If a deadend is correct, then any FSAP that covers an action leading to it is also correct. If all actions executable in a state are forbidden by correctly identified FSAPs, then that state is itself a deadend, and the process proceeds backwards.

In the final round, the FSAP avoidance will mean the search for weak plans will only be in a space where strong cyclic solutions exist (otherwise we would have a contradiction in this being the final round). Since weak plans will always be found to build the policy, every node reached will have some solution found and included. Because of the exhaustive nature of processing nodes in \csgraph, eventually, a \psgraph will be computed such that it is detected as being strong cyclic.
}

\section{Evaluation}
\label{sec:evaluation}

Our objective is to understand the performance of \us compared to other FOND planners in terms of coverage, solution size, and solve time.
We implemented \us on top of the Fast Downward planning system \cite{fd} and used very little of the released code for PRP: specifically, much of the translation and scripts were re-used, as is the case with other FOND planners (e.g., MyND and FONDSAT use the same parsing mechanism). The code, benchmarks, and detailed analysis can be found at \verb|mulab.ai/pr2| .

We compare against the state of the art in FOND planning: MyND \cite{mynd}, FONDSAT \cite{fondsat}, PRP \cite{prp}, and Paladinus \cite{paladinus}. We configured each planner to its best settings based on aggregate performance across all domains, including using a modern SAT solver for FONDSAT (improving its coverage by a fair margin).
Planners were given 4Gb of memory and 60min to solve an instance, and evaluations were run on a PowerEdge C6420 machine running Ubuntu with an Intel 5218 2.3GHz processor.

To evaluate our planners, we collected all of the benchmarks employed for evaluation of the FOND planners listed above, representing a total of 18 domains.
Across the 18 domains, there is a wide disparity in the number of instances, from 8 in the smallest (acrobatics) to 190 in the largest (faults-new). Consequently, we normalize the coverage on a per-domain basis to be a maximum of~1.

\subsection{Planner Comparison}

In Table \ref{tbl:coverage} we show the normalized coverage across all domains and planners. \us performs at least as well, and often better, than every other planner in virtually every domain. The one exception is for blockworlds-new, where the problem size for a single instance grows to an extent that \us runs out of memory while PRP just barely does not (it is the largest instance PRP is capable of solving). Additionally, there are four instances in forest-new that PRP solves and \us does not (though, several other instances that \us solves and PRP does not). Not only does \us perform well on the older benchmarks where PRP was known to be state of the art, but it also handles every one of the new benchmark domains -- islands, miner, tire-spiky, and tire-truck.

We should note that in the domain `doors' for PRP (cf. (*) in Table \ref{tbl:coverage}), a bug in PRP led it to incorrectly declare the three simplest problems have no solution. Presumably, if this bug were fixed, the performance would increase by 0.2 to match \us and FONDSAT with a perfect score. We have removed the instances (7 from firstresponders-new and 3 from the original tireworld) for which no strong cyclic solutions exist; while PR2 correctly returns that no strong cyclic solutions exist for these instances, some other planners fail in unpredictable ways. We did not detect any further erroneous behaviour among the planners and domains.

\begin{table}[t]
\small
\centering
\begin{tabular}{lrrrrr}
\hline
 domain (size)    &   pr2 &   prp &   fsat &   pala &   mynd \\
\hline
 acrobatics (8)   & \textbf{1.00}  & \textbf{1.00} & 0.50          & \textbf{1.00} & \textbf{1.00} \\
 beam-walk (11)   & \textbf{1.00}  & \textbf{1.00} & 0.27          & 0.82          & \textbf{1.00} \\
 bw-new (50)      & 0.82           & \textbf{0.84} & 0.16          & 0.36          & 0.42          \\
 chain (10)       & \textbf{1.00}  & \textbf{1.00} & 0.10          & \textbf{1.00} & \textbf{1.00} \\
 earth-obs (40)   & \textbf{1.00}  & \textbf{1.00} & 0.17          & 0.65          & 0.78          \\
 elevators (15)   & \textbf{1.00}  & \textbf{1.00} & 0.47          & 0.53          & 0.93          \\
 faults-new (190) & \textbf{1.00}  & \textbf{1.00} & 0.04          & 0.12          & 0.84          \\
 first-new (88)   & \textbf{0.99}  & \textbf{0.99} & 0.06          & 0.06          & 0.09          \\
 forest-new (100) & \textbf{0.94}  & 0.88          & 0.10          & 0.18          & 0.16          \\
 tidyup-mdp (10)  & \textbf{1.00}  & 0.00          & 0.10          & 0.40          & \textbf{1.00} \\
 tire (12)        & \textbf{1.00}  & \textbf{1.00} & \textbf{1.00} & \textbf{1.00} & \textbf{1.00} \\
 tri-tire (40)    & \textbf{1.00}  & \textbf{1.00} & 0.10          & 0.20          & 0.17          \\
 zeno (15)        & \textbf{1.00}  & \textbf{1.00} & 0.33          & 0.53          & 0.60          \\
 doors (15)       & \textbf{1.00}  & (*) 0.80          & \textbf{1.00} & 0.93          & 0.73          \\
 islands (60)     & \textbf{1.00}  & 0.52          & 0.95          & \textbf{1.00} & 0.22          \\
 miner (51)       & \textbf{1.00}  & 0.25          & 0.98          & \textbf{1.00} & 0.00          \\
 tire-spiky (11)  & \textbf{1.00}  & 0.09          & 0.91          & 0.91          & 0.18          \\
 tire-truck (74)  & \textbf{1.00}  & 0.28          & 0.99          & 0.65          & 0.20          \\
 TOTAL (800)      & \textbf{17.75} & 13.65         & 8.23          & 11.34         & 10.33    \\
\hline
\end{tabular}
\caption{Normalized Coverage for all planners and domains.}
\label{tbl:coverage}
\end{table}

\begin{figure*}[t]
    \centering
    \begin{subfigure}[b]{0.31\textwidth}
        \includegraphics[width=\textwidth]{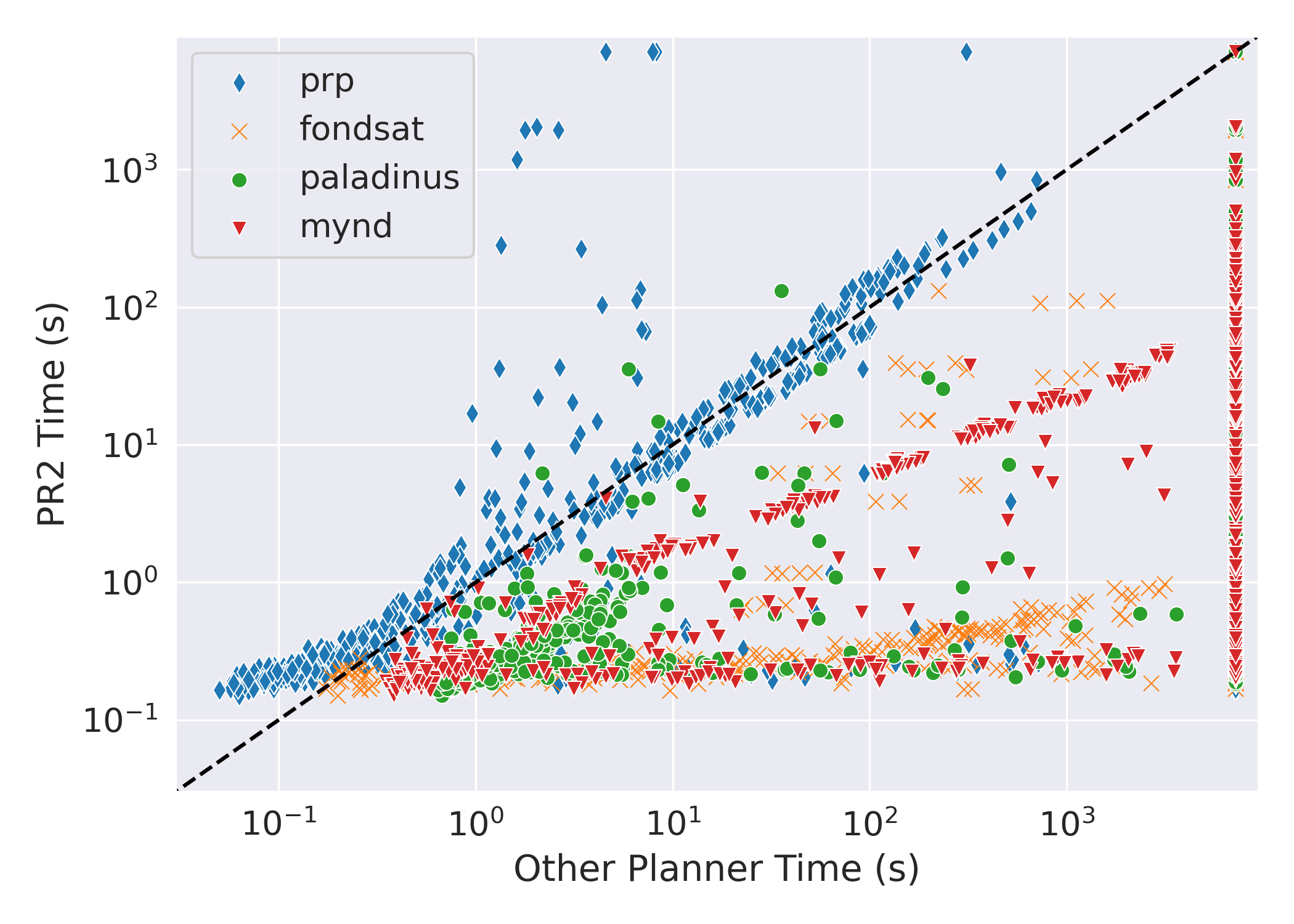}
        \caption{Time comparison.}
        \label{fig:time-comparison}
    \end{subfigure}
    \hfill
    \begin{subfigure}[b]{0.31\textwidth}
        \includegraphics[width=\textwidth]{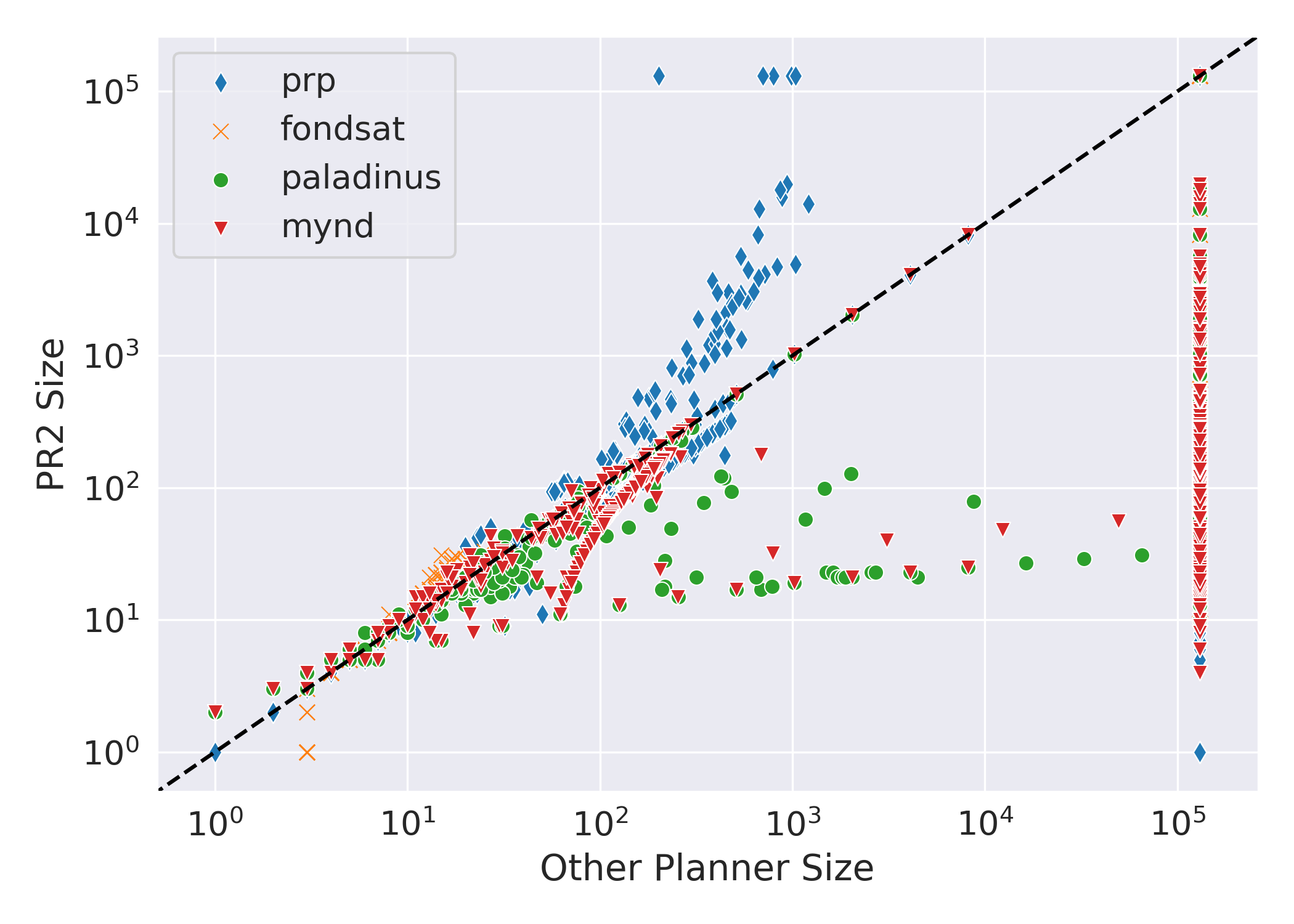}
        \caption{Size comparison.}
        \label{fig:size-comparison}
    \end{subfigure}
    \hfill
    \begin{subfigure}[b]{0.37\textwidth}
        \includegraphics[width=\textwidth]{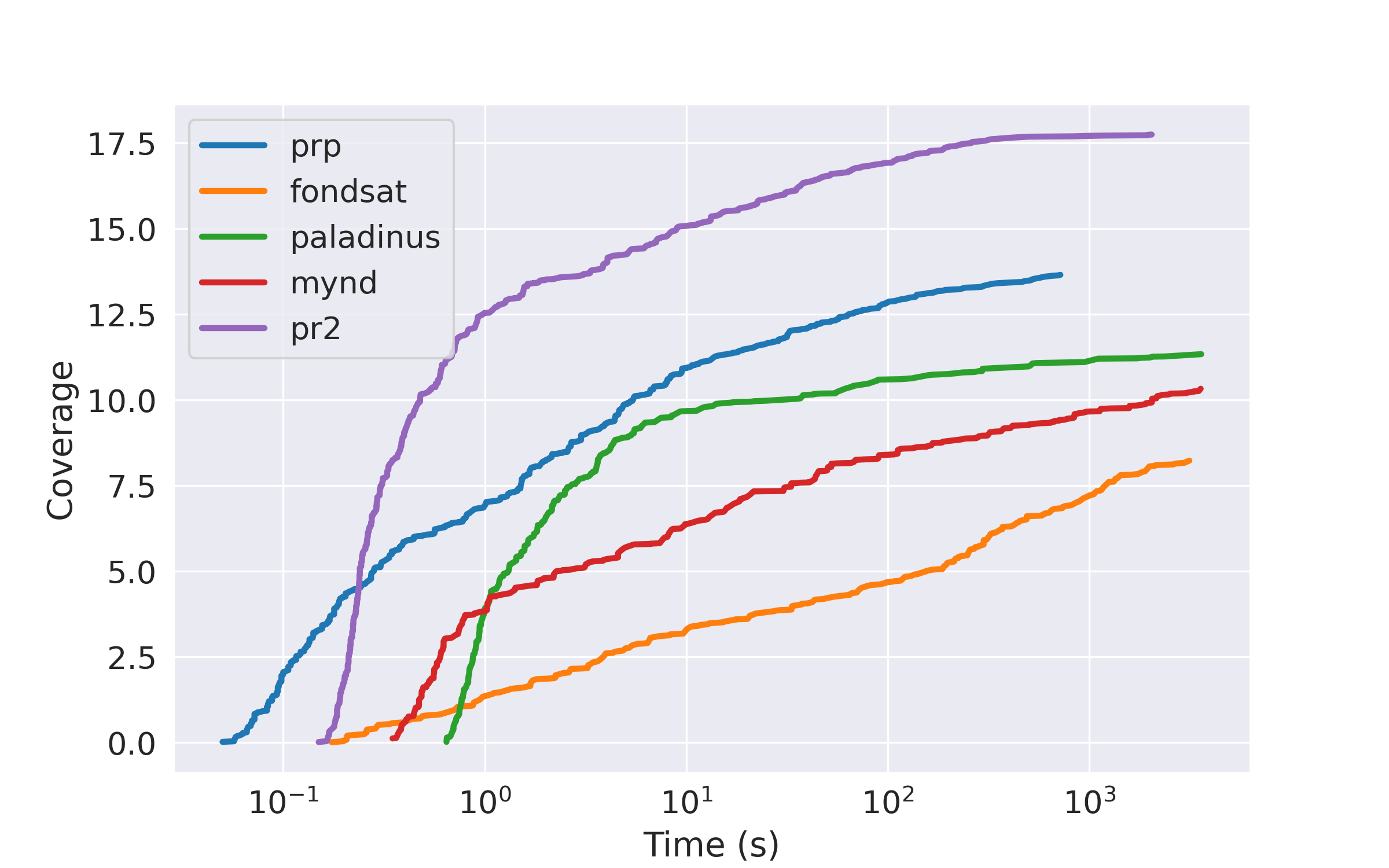}
        \caption{Coverage over time.}
        \label{fig:coverage}
    \end{subfigure}
    \caption{Comparison of time (left), size (middle), and coverage (right). \textbf{Time:} a log-log plot showing the \us runtime against the other planners, over all problems. Anything below the x=y line is an improvement. \textbf{Size:} similar to time, but measured in the solution size produced by each planner. \textbf{Coverage:} ``Survival plot'' that shows the normalized coverage achieved as a function of time for each of the planners. Note the log scaling on the x-axis.}
    \label{fig:comparison}
\end{figure*}

We report on the time and solution size comparison in Figures \ref{fig:time-comparison} and \ref{fig:size-comparison}.
\us consistently outperforms the other planners on both measures. However, PRP produces smaller solutions as the problem size grows (the PRP planner uses a solution representation not shared by any of the other planners), and is also faster in a handful of problems.
Finally, in Figure \ref{fig:coverage}, we show the normalized coverage over time to give a sense of how quickly the solutions are found. After roughly 0.5 seconds, \us surpasses PRP in terms of coverage and remains dominant.

\subsection{Ablation Studies}
\label{sec:ablation-evals}

We briefly report on the performance impact of several of the features of \us. Corresponding to Sections \ref{sec:scm} - \ref{sec:1safe}, we disabled each feature and measured the impact on normalized coverage in the \us planner:
\vspace{0.5em}

\begin{tabular}{lc}
\multicolumn{1}{c}{(section) \textbf{Disabled Feature}}     & \textbf{Drop in coverage} \\[0.5em]
(\ref{sec:scm}) Full Strong-Cyclic Marking & -0.757                 \\
(\ref{sec:deadends}) Poisoning             & -0.743                 \\
(\ref{sec:heuristic}) FSAP Heuristic       & -2.839                 \\
(\ref{sec:obj-sampling}) Object Sampling   & -0.743                 \\
(\ref{sec:1safe}) Forcing 1-safe Plans     & -1.570                
\end{tabular}

\vspace{0.5em}
First, we note that no instance unsolved by \us was solved by disabling one of the features. Thus, their presence never hurts the planner in terms of coverage. To place the reduction of coverage in perspective, a drop of 1.0 indicates (roughly) a full domain's worth of instances that cannot be solved. In the supplementary material, we provide a full table similar to Table \ref{tbl:coverage}, but the key takeaways are: (1) removing any one of the features drops the coverage in the \textit{tireworld-truck} domain substantially (from nearly 1.0 down to 0.2-0.3); (2) the drop in coverage for poisoning, object sampling, and full strong-cyclic marking is entirely due to this one domain; (3) disabling the FSAP heuristic and forcing 1-safe plans reduced performance greatly in the \textit{triangle tireworld} domain; and (4) disabling the FSAP heuristic degraded performance in the \textit{islands} and \textit{miner} domains.

We can conclude that every one of the introduced features provides a net benefit to the planner, which (at times) is crucial to achieving peak performance. Moreover, the FSAP-aware heuristic appears to have the biggest impact on the results of the \us planner. We note that removing any single feature leaves \us with a planner that still outperforms all previous FOND planners.

\section{Related Work}
\label{sec:related}

The closest related work is the FOND planner, PRP \cite{prp}. \us falls under the same category of FOND planner approach as PRP, along with FIP and NDP \cite{fip,ndp}. The strategy \us uses to leverage regression is similar to PRP, including the handling of conditional effects \cite{primf}, while many of the generalization methods are shared across a broader set of planners (e.g., the deadend relaxation introduced by SixthSense \cite{KolobovMW10}).
Concretely, very little implementation is shared between PRP and \us. Beyond the implementation, the solution representation (interplay between \psgraph and \csgraph) and the algorithms surrounding them are also novel. From the custom FSAP-aware heuristic to the procedures for strengthening and marking \psgraph, all techniques detailed in Sections \ref{sec:fpr} - \ref{sec:1safe} are novel.

The \psgraph representation is shared by the FONDSAT and GRENDEL planners \cite{fondsat,grendel}. Having a controller with regressed conditions associated with the nodes strikes a powerful balance between representation size and generality. Nonetheless, the three planners (\us, FONDSAT, and GRENDEL) take strikingly different approaches to computing the solution. Whereas \us uses the replanning approach pioneered by NDP and extended by FIP and PRP, FONDSAT uses a SAT encoding of a structure very close to the \psgraph,
and GRENDEL uses backwards search through the space of controllers. Finally, there is a similarity between the fixed-point regression performed by \us on the \psgraph and GRENDEL's approach to re-establish the conditions needed for strong-cyclic solutions.

The final category of FOND solvers is the policy-space search of MyND and Paladinus \cite{mynd,paladinus}. By many measures, Paladinus represents the state of the art for this class of solution strategy. Particularly on the newer domains, it outperforms MyND by a large margin. These planners operate by searching through the space of reachable policies, incrementally adding to them until a strong cyclic policy is found.

\section{Concluding Remarks}
\label{sec:conclusion}

The release of PRP in 2012 marked a significant jump in our ability to solve FOND problems, opening the door to new application areas, and inspiring the development of contingent~\cite{contingentprp}, probabilistic~\cite{probprp} and LTL-FOND~\cite{camacho-ltl} variants. We present a new FOND planner, \us, that similarly advances the state of the art in FOND planning. The \us planner was designed from the ground up using many of the best-known techniques for FOND planning. We introduced a suite of novel techniques that help \us achieve its high performance, and the planner itself is amenable to extension and further research.

A promising direction for such work is in the further development of novel heuristics for FOND planning. The FSAP-aware heuristic proposed here (cf. Section \ref{sec:heuristic}) is a significant improvement over previous methods, but it is only one idea in the arena of FOND-aware heuristics. From a more theoretical viewpoint, there is a growing recognition that the solution representation of a FOND problem plays a critical role in computing and executing plans \cite{victoria-intex,icaps-fond-solutions}. The merits of the representational power of the \psgraph\ as compared to the partial state-action pairs of PRP or the complete state-action mappings found in other FOND planners, deserves further study. Finally, our results show that FOND planning has achieved a notable leap in performance. New benchmark problems, perhaps from adjacent fields such as reactive synthesis \cite{gdg-vardi-synthesis15,camacho-ltl}, should be assembled in service of a modern suite of challenge domains, afforded by these new capabilities.

\subsection*{Acknowledgements}
We gratefully acknowledge funding from the Natural Sciences and Engineering Research Council of Canada (NSERC), the Vector Institute for Artificial Intelligence and the Canada CIFAR AI Chair program.

    \bibliography{references}

\ifthenelse{\equal{\compileaaai}{}}%
{

    \onecolumn
    \newpage
    \twocolumn

\appendix

\section{Full High-Level Algorithm}
\label{app:high-level}

In Algorithm \ref{alg:high-level-full} we give a more complete picture of the high-level procedure summarized in Algorithm \ref{alg:high-level}.

\begin{algorithm}[ht]
\KwIn{FOND planning task, $\Pi = \fondproblem{s_0}{s_*}$}
\KwOut{Partial policy P}
incumbent = \texttt{make\_empty\_solution}()\;
FSAPS = $\emptyset$\;

\While{!incumbent.is\_strong\_cyclic()}
{
  P = \texttt{make\_empty\_solution}()\;
  open\_list = $\set{\texttt{make\_node}(s_0)}$\;
  \While{!open\_list.empty()}
  {
    n = open\_list.pop()\;
    \tcp{Case 0: Node is handled by \psgraph}
    handled = $P.\psgraph(n).sc$\;
    \tcp{Case 1: Node is poisoned}
    \If {!handled}
    {
        handled = \texttt{check\_poisoned}(n, P)\;
    }
    \tcp{Case 2: Complete State Match}
    \If {!handled}
    {
      handled = \texttt{match\_complete\_state}(n, P)\;
    }
    \tcp{Case 3: Pre-defined Path}
    \If {!handled}
    {
      handled = \texttt{predefined\_path}(n, P)\;
    }
    \tcp{Case 4: Hookup Solution Steps}
    \If {!handled}
    {
      handled = \texttt{match\_complete\_state}(n, P)\;
    }
    \tcp{Case 5: New Weak Plan}
    \If {!handled}
    {
      handled, $\pi$ = \texttt{find\_weak\_plan}(n, $\Pi$, P, FSAPS)\;
      \If {handled}
      {
        \tcp{Add new solution fragment and expand openlist}
        \texttt{update\_plan}(n, $\pi$, P, open\_list)\;
      }
    }
    \tcp{Case 6: Deadend State}
    \If {!handled}
    {
      \texttt{record\_deadend}(n, FSAPS)\;
      \If {n.state == $s_0$}
      {
        \Return{incumbent}\;
      }
    }
  }
  \If {P.success\_rate() $\geq$ incumbent.success\_rate()}
  {
    incumbent = P\;
  }
}

\Return{incumbent}\;
\caption{Expanded \us High-Level Planner}
\label{alg:high-level-full}
\end{algorithm}

\section{FSAP-Aware FF Heuristic}
\label{app:heuristic}

Here, we focus on the core modifications made to the FF heuristic and refer the interested reader to the FF and FD planning systems for a more systematic treatment of the original heuristic mechanics \cite{ff,fd}. There are two key changes made to the FF heuristic to take the FSAPs into account. Firstly, when computing preferred operators for a state in the search, we only allow those that are not forbidden. The preferred operator may stem from an outcome that was not the source of the existing FSAP (i.e., the deadend was found down a different non-deterministic outcome), so we must forgo using it as a preferred operator (aka. helpful action).

The second key change is to the heuristic value itself. Intuitively, we add a penalty to any operator that is used in the computation of the heuristic value, which \textit{may} be forbidden. This ``potentially forbidden'' property is calculated by observing which FSAPs have their conditions satisfied by the propositions reachable in the delete relaxation. To simplify the exposition, we use $\props{\vars}$ to refer to all the variable-value pairs that are possible. The modification is detailed in Algorithms \ref{alg:heuristic} and \ref{alg:heur-cost}.

\begin{algorithm}[ht]
\KwIn{State $s$, planning problem $\fondproblem{s_0}{s_*}$}
\KwOut{Heuristic estimate for $s$}

\vspace{0.5em}
\tcp{Seed every proposition and action with initial costs.}
Q = []\;
\For{$prop \in \props{\vars}$}
{
    \If{$s(prop.var) == prop.val$}
    {
        Q.push\_back($\langle prop, 0 \rangle$)\;
        $prop$.heur = 0\;
    }
    \Else{
        $prop$.heur = $\infty$\;
    }
}
\For{$act \in \act$}
{
    $act$.heur = 0\;
    $act$.prec\_left = $|\pre{act}|$\;
}

\vspace{0.5em}
\tcp{Process the queue until empty}
\While{Q is not empty}
{
    $\langle prop, cost \rangle$ = Q.pop(0)\;

    \vspace{0.5em}
    \tcp{Loop over the actions that use this proposition}
    \For{$act \in \set{a \st a \in \act \text{ and } prop \in \pre{a}}$}
    {
        \vspace{0.5em}
        \tcp{Increase the cost and check if all preconditions are satisfied}
        $act$.heur += cost\;
        $act$.prec\_left -= 1\;
        \If{$act$.prec\_left == 0}
        {
            \vspace{0.5em}
            \tcp{Add a penalty for potential FSAPs}
            $new\_cost = act$.heur + \textbf{\color{blue}\texttt{FSAPCost}($act$)}\;

            \vspace{0.5em}
            \tcp{Queue the effects of this action with updated cost}
            \For{$prop' \in \eff{act}$} {
                \If{$prop'$.heur $ > new\_cost$} {
                    $prop'$.heur = $new\_cost$\;
                    Q.push\_back($\langle prop', new\_cost \rangle$)\;
                }
            }
        }
    }
}

\vspace{0.5em}
\tcp{Compute the total goal cost.}
$total = 0$\;
\For{$prop \in s_*$}
{$total$ += $prop$.heur\;}
\Return{$total$}\;

\caption{FSAP-aware FF Heuristic}
\label{alg:heuristic}
\end{algorithm}

\begin{algorithm}[ht]
\KwIn{Action $act$, FSAP set FSAPS, and a constant penalty cost $C$}
\KwOut{Adjusted cost for action $act$}

\vspace{0.5em}
\tcp{Keep track of the FSAPs we see.}
$Seen = \emptyset$\;

\vspace{0.5em}
\tcp{Loop over all FSAPs from the search that match $act$.}

$total = act$.cost\;
\For{$\mathit{fsap} \in \mathrm{FSAPS} \hspace{0.5em}\mathrm{s.t.,}\hspace{0.5em} \mathit{fsap}.a == act$}
{
    \vspace{0.5em}
    \tcp{Just increase the cost if it's already seen.}
    \If{$\mathit{fsap} \in Seen$}
    {
        $total$ += $C$\;
    }
    \vspace{0.5em}
    \tcp{If the FSAP may have triggered, add it.}
    \Else{
        \If{$\forall \hspace{0.3em} prop \in \mathit{fsap}.ps, \hspace{0.3em} prop$.heur $ < \infty$}
        {
            $Seen$.add($\mathit{fsap}$)\;
            $total$ += $C$\;
        }
    }
}
\Return{$total$}\;
\caption{\texttt{FSAPCost} Algorithm}
\label{alg:heur-cost}
\end{algorithm}

Note that much of Algorithm \ref{alg:heuristic} follows a standard implementation of a delete-relaxed classical planning heuristic, with the exception of dynamically adjusting the action costs (via Algorithm \ref{alg:heur-cost}) based on potential FSAP violations. The correctness of the algorithm is preserved as long as the constant cost $C$ is finite (the heuristic remains inadmissible).

\section{Redundant Object Sampling}
\label{app:object-sampling}

Prior to attempting to solve a problem, \us will explore simplifications to the instance that greatly reduce its complexity by attempting to remove redundant objects. This sound-but-incomplete transformation comes from the following observation: \textit{if we delete some of the objects in a problem, and the resulting instance has a strong cyclic solution $sol$, then $sol$ must be a solution to the original problem.} Intuitively, our removal of objects has an impact equivalent to disallowing or deleting several actions, thus not invalidating any solutions that remain.

To ``delete'', we remove the object and any fluent referring to it from the problem file (we forgo doing this for constants for now). To generate a set of candidate objects to remove, we partition all objects into symmetric sets: two objects are considered symmetric if they appear in exactly the same way across the problem file (i.e., they appear in similar fluents in the exact same way). No object is eligible for removal if it appears in the goal. This is an approximation of a more general form of object symmetries, but it nonetheless provides several compelling sets of symmetric objects.

With the categories calculated, we sample a subset of the objects from each symmetric set and try to solve the problem with limited computational resources. The schedule of object sampling and time limits given were computed using a general sampling of some of the existing FOND domains and default to the following (attempted from top to bottom):

\begin{center}
\begin{tabular}{|c|c|}
\hline
\# objects used & Time Limit (sec) \\
\hline
1 & 60 \\
2 & 240 \\
4 & 30 \\
8 & 30 \\
$\infty$ & 3240 \\
\hline
\end{tabular}
\end{center}

Note that the final value of 3,240 seconds reflects the remaining time from a 60min timeout after the first four configurations fail to find a solution. The planner stops as soon as a solution is found for any configuration.

\section{Detailed Ablation Results}
\label{app:full-ablation-results}

In Table \ref{tbl:full-ablation}, we detail the per-domain (normalized) coverage for each variant of \us that was run (disabling one of the features, as detailed in Section \ref{sec:ablation-evals}). The summarized results in Section \ref{sec:ablation-evals} come from the difference between an ablation column and the first \us column (i.e., full planner).

\begin{table*}[t]
\small
\centering
\begin{tabular}{lcccccc}
\hline
 domain (size)    &   pr2 &   no-objsampling &   no-poisoning &   no-fsap-penalty &   no-full-scd-marking &   no-force-1safe \\
\hline
 acrobatics (8)   &  1.00 &                 1.00 &               1.00 &                  1.00 &                      1.00 &                 1.00 \\
 beam-walk (11)   &  1.00 &                 1.00 &               1.00 &                  1.00 &                      1.00 &                 1.00 \\
 bw-new (50)      &  0.82 &                 0.82 &               0.82 &                  0.82 &                      0.82 &                 0.82 \\
 chain (10)       &  1.00 &                 1.00 &               1.00 &                  1.00 &                      1.00 &                 1.00 \\
 earth-obs (40)   &  1.00 &                 1.00 &               1.00 &                  1.00 &                      1.00 &                 1.00 \\
 elevators (15)   &  1.00 &                 1.00 &               1.00 &                  1.00 &                      1.00 &                 1.00 \\
 faults-new (190) &  1.00 &                 1.00 &               1.00 &                  1.00 &                      1.00 &                 1.00 \\
 first-new (88)   &  0.99 &                 0.99 &               0.99 &                  0.99 &                      0.99 &                 0.99 \\
 forest-new (100) &  0.94 &                 0.94 &               0.94 &                  0.94 &                      0.94 &                 0.94 \\
 tidyup-mdp (10)  &  1.00 &                 1.00 &               1.00 &                  1.00 &                      1.00 &                 1.00 \\
 tire (12)        &  1.00 &                 1.00 &               1.00 &                  1.00 &                      1.00 &                 1.00 \\
 tri-tire (40)    &  1.00 &                 1.00 &               1.00 &                  0.28 &                      1.00 &                 0.20 \\
 zeno (15)        &  1.00 &                 1.00 &               1.00 &                  1.00 &                      1.00 &                 1.00 \\
 doors (15)       &  1.00 &                 1.00 &               1.00 &                  1.00 &                      1.00 &                 1.00 \\
 islands (60)     &  1.00 &                 1.00 &               1.00 &                  0.52 &                      1.00 &                 1.00 \\
 miner (51)       &  1.00 &                 1.00 &               1.00 &                  0.06 &                      1.00 &                 1.00 \\
 tire-spiky (11)  &  1.00 &                 1.00 &               1.00 &                  1.00 &                      1.00 &                 1.00 \\
 tire-truck (74)  &  0.99 &                 0.24 &               0.24 &                  0.30 &                      0.23 &                 0.22 \\
 TOTAL (800)      & 17.74 &                16.99 &              16.99 &                 14.90 &                     16.98 &                16.16 \\
\hline
\end{tabular}
\caption{Normalized Coverage for all planners and domains.}
\label{tbl:full-ablation}
\end{table*}

}{} 

\end{document}